\documentclass[11pt]{article}

\usepackage{microtype} % microtypography
\usepackage{booktabs}  % tables
\usepackage{url}  % url

\usepackage{amsmath}
\usepackage{amsthm}

\usepackage[final]{automl}
\usepackage{natbib}
\usepackage{graphicx}
\usepackage{float}
\newcommand*{\hmm}[1]{#1\nobreak\discretionary{}{\hbox{$\mathsurround=0pt #1$}}{}}

\title{Layer-Specific Optimization: Sensitivity Based Convolution Layers Basis Search}

\author[1]{\nameemail{Vasiliy Alekseev}{vasiliy.alekseyev@phystech.edu}}
\author[1]{\nameemail{Ilya Lukashevich}{lukashevich.ia@phystech.edu}}
\author[1]{\nameemail{Ilia Zharikov}{ilya.zharikov@phystech.edu}}  % ilya.zharikov@phystech.edu zharikov.i.n@yandex.ru
\author[1]{\nameemail{Ilya Vasiliev}{ilya.v.vasilev@phystech.edu}}

\affil[1]{Moscow Institute of Physics and Technology, Moscow, Russia}

\hypersetup{%
  pdfauthor={}, % will be reset to "Anonymous" unless the "final" package option is given
  pdftitle={},
  pdfsubject={},
  pdfkeywords={}
}

\begin{document}

\maketitle

\begin{abstract}
    Deep neural network models have a complex architecture and are overparameterized. The number of parameters is more than the whole dataset, which is highly resource-consuming. This complicates their application and limits its usage on different devices.
    Reduction in the number of network parameters helps to reduce the size of the model, but at the same time, thoughtlessly applied, can lead to a deterioration in the quality of the network.
    One way to reduce the number of model parameters is matrix decomposition, where a matrix is represented as a product of smaller matrices.
    In this paper, we propose a new way of applying the matrix decomposition with respect to the weights of convolutional layers.
    The essence of the method is to train not all convolutions, but only the subset of convolutions (\emph{basis convolutions}), and represent the rest as linear combinations of the basis ones.
    Experiments on models from the ResNet family and the CIFAR-$10$ dataset demonstrate that basis convolutions can not only reduce the size of the model but also accelerate the forward and backward passes of the network.
    Another contribution of this work is that we propose a fast method for selecting a subset of network layers in which the use of matrix decomposition does not degrade the quality of the final model.
\end{abstract}

\section{Introduction}

% TODO

% \paragraph{О чём.}
% Уменьшение числа параметров в модели.
% Ускорение обучения и/или inference.
% Уменьшение размера модели (квантизация, прунинг, KD, ...?).

% \paragraph{\textcolor{cian}{Главное в работе.}}
% \begin{itemize}
%     \item Представлен способ обучения сети с помощью \emph{базисных свёрток}. Проведено сравнение с другими способами декомпозиции: TTD, Tucker. Показано, что предлагаемый способ, помимо уменьшения размера модели, позволяет получать выигрыш в скорости обучения сети.
%     \item Предложен способ выборочного применения метода декомпозиции (базисные свёртки) к конкретным слоям сети, с тем чтобы итоговое качество уменьшалось не сильно по сравнению с оригинальной сетью. Проведены эксперименты на моделях семейства ResNet. 
% \end{itemize}

Deep learning has gained significant attention in recent years due to the promising results in various scientific fields.
However, a real-world application of the developed models and methods remains a complicated problem.
Modern neural network architectures suffer from a problem of parameter redundancy resulting in large memory consumption and high training and/or inference time.
The utilization of created approaches and their deployment on resource-constrained devices is limited due to the existing difficulties.
Therefore, neural network compression methods are created for solving the problem.

A series of compression methods are focused on the decomposition of weight matrices or tensors.
The motivation behind the idea is based on the observation of a low-rank structure of fully connected or convolutional layers~\citep{Denil2013PredictingPI, Jaderberg2014SpeedingUC}.
To perform matrix decomposition of weights, the tensors are generally folded into $2$-dimensional ones across the specified dimensions.
Thereafter, this representation is decomposed into a set of matrices in which the total number of parameters is significantly lower than in the original tensor.
The idea of the simultaneous low-rank structure and sparsity in weight matrices is also considered~\citep{8099498}.
Under such assumptions, the tensor can be decomposed into the sum of the product of two low-rank matrices and a sparse matrix.
However, a sparse matrix can possibly induce additional costs in terms of computational operations and reduce the effectiveness of the method.
In order to exploit the multidimensional structure of neural network weights, the tensor decomposition is applied~\citep{Kim2016CompressionOD, Astrid2017CPdecompositionWT}.
With the application of a tensor-based set of techniques, a greater reduction in the number of network parameters can be achieved.

A significant proportion of research works is focused on decomposing the pre-trained weights of the model with a following fine-tuning stage.
The described procedure does not concentrate on training time optimization.
On the contrary, another part of the methods considers trainable low-rank approximations of tensors that are learned throughout training~\citep{garipov2016ultimate}.
The latter approach allows achieving higher network compression rates while potentially decreasing both training and inference time, which is one of the goals of the present work. 

To reduce the training time and the number of neural network parameters, a method for training the basis convolutions is proposed.
Different setups of the forward propagation using the basis convolutions and composition coefficients are considered for reaching higher training acceleration.
Moreover, a search method for an optimal configuration of training with basis convolutions is introduced.
The approach is based on the sensitivity analysis of network layers to decomposition and allows discovering settings with the least possible decrease in the final model performance in comparison to the original model.
The proposed methods were tested on ResNet architectures and compared with other compression methods such as Tensor Train Decomposition.

% As a result, to train models with fewer parameters, we did the following:
% \begin{itemize}
%     \item proposed convolutional layers' forward pass modification and reduced the coefficients matrix which allowed to restore the outputs of the convolutions based on the outputs of basis convolutions;
%     \item proposed an approach to estimate sensitivity of layer groups to training with basis convolutions;
%     \item proposed an approach of finding an optimal layer combination to train with basis convolutions in terms of training time and resulting accuracy;
%     \item validated the proposed approach and showed that basis convolutions can help to train a model faster without a drop in final accuracy.
% \end{itemize}

\section{Related Work}

% \paragraph{О чём.}
% CNN.
% TTD (в частности, \citep{taskynov2021tensor}).
% Tucker decomposition.
% ZeroQ \citep{cai2020zeroq} как способ выбора частей сети, которые можно ``оптимизировать'' с минимальной потерей качества (поиск оптимальной mixed precision конфигурации квантизации сети).

Neural network compression methods based on the decomposition of weights can be divided into two major groups: matrix-based methods and tensor-based ones.

\paragraph{Matrix decomposition} 
Originally, \citet{Denil2013PredictingPI} exploited the low-rank property of the tensors folded into $2$-dimensional matrices for the parameter prediction task.
\citet{10.5555/2968826.2968968} combined a singular value decomposition (SVD) with filter clustering, which resulted in $2\times$ compression with a $1\%$ drop in accuracy.
\citet{tai2016convolutional} proposed a new algorithm for a low-rank tensor decomposition based on the SVD and achieved $3-5 \times$ weight reduction without a significant drop in accuracy for a set of CNN architectures.
To employ the low-rank and sparsity features of neural network weights, \citet{8099498} represented the weight matrix $W$ as $W \approx L + S$, where $L$ is a low-rank component and $S$ is a sparse matrix.
Combining the decomposition with an asymmetric data reconstruction term resulted in $15\times$ reduction in model size for a VGG-$16$ model.
The sparse patterns in the decomposed low-rank approximations of weight matrices were explored by \citet{8962461} to get high compression ratios and to mitigate the performance degradation.
A specific modern feature of utilizing the common components in different layers of a neural network has been studied.
The main goal of the methods considering this characteristic is to decompose the original tensors into smaller ones, part of which is common for all model weights and the remaining part is specific for each particular weight.
\citet{chen2021joint} proposed three joint decomposition schemes based on the SVD exploit the idea and achieved $22\times$ reduction in model size with a lower drop in accuracy for a ResNet34 model compared with the other methods. 

\paragraph{Tensor decomposition}
The weight tensors in neural networks are usually multidimensional.
In order to utilize the tensor-based structure, the tensor decomposition methods are developed.
\citet{Kim2016CompressionOD} proposed to apply a Tucker decomposition on the kernel tensors.
The rank of the decomposition is determined by the solution of variational Bayesian matrix factorization \citep{NIPS2012_26337353}.
A low-rank CP-decomposition scheme with Tensor Power Method was implemented by \citet{Astrid2017CPdecompositionWT} to achieve higher compression ratios and speed-ups compared to the Tucker decomposition.
\citet{7449822} employed Kronecker product decomposition to compress the fully-connected layers of networks by replacing the weight tensor with a linear combination of smaller tensors represented by the Kronecker products.
\citet{garipov2016ultimate} proposed reshaping method of $4$-dimensional kernel of the convolution into a multidimensional tensor and applied Tensor Train Decomposition (TTD) to it.
Combining the described approach with a fully-connected layer compression method derived by \citet{NIPS2015_6855456e}, the maximum $82\times$ compression with a $1\%$ drop in model performance was obtained.
\citet{taskynov2021tensor} presented a one-shot training algorithm based on the TTD for accelerating models which takes into account the order of decomposition.
The method is focused on reducing the latency on specific hardware devices like Ascend 310 NPU.
This technique allows speeding up the ResNet architectures by approximately $15\%$ with a drop of $0.1\%$ in final top-$1$ accuracy on ImageNet.
Additionally, the common structure of the layers in neural network architectures is explored via tensor decomposition methods.
The fully and partly coupled TTD were proposed by \citet{9261106} to identify the common and specific information in weight tensors, which resulted in higher performance for ResNet models under the same compression ratios compared with the SVD and usual TTD.

\paragraph{Optimal configuration search}
In order to effectively apply compression or acceleration to the neural networks, we need to derive an optimal configuration of layers where the methods are applied.
A naive searching approach for the best setting requires an exponential number of training runs.
To alleviate the similar problem, \citet{cai2020zeroq} proposed a method for an automatic search for a mixed-precision configuration of bit-widths in neural network quantization problem.
The authors introduced a Pareto frontier-based approach, which decreases the computational overhead of the search from exponential to linear in the number of layers.
The suggested method enabled the authors to achieve $1.72\%$ higher accuracy for a MobileNetV2 model than its counterparts in the quantization task with lower expenses. 

% \section{CNN}

% \paragraph{О чём.}
% Архитектура CNN сетей.
% Принцип работы Conv слоя.Задачи, в которых применяются CNN (?).

% \paragraph{Что будет.} Иллюстрация работы Conv слоя (далее в секции про базисные свёртки будут похожие по формату иллюстрации).

\section{Method}

\subsection{Basis Convolutions}
\label{sec:basis_convolutions}

% \paragraph{О чём.}
% Несколько способов обучать со свёртками (замена weight, модификация forward).

% \paragraph{Что будет.}
% Иллюстрации для подходов.
% Теоретическая оценка ускорения (числа операций) на стадиях forward и backward.

% \paragraph{Вывод.}
% Ожидается ускорение обучения при применении базисных свёрток (модификация forward).

% \bigskip

The core idea behind using decomposition is to keep just a subset of trainable model parameters.
Hypothetically, the lower the number of parameters, the faster the training.
However, the resulting quality may deteriorate.
We believe that there is a possibility to reduce the number of trainable parameters without a significant drop in quality.

\begin{figure}[h]  % TODO: svg, not png
    \centering
    \includegraphics[width=0.8\textwidth]{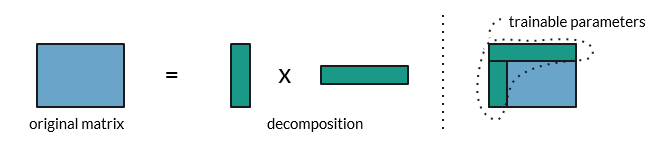}
    \caption{Matrix decomposition.}
    \label{fig:matrix_decomposition_illustration}
\end{figure}

There are several approaches which can be used to reduce the number of trainable parameters in the weight matrices of $2$D convolution modules (Conv2d).
First, matrix decomposition.
For example, QR or SVD~(see Fig.~\ref{fig:matrix_decomposition_illustration}).
This is a $2$-dimensional approach, which means that the weight matrix is supposed to be $2$-dimensional.

Another way to reduce the number of parameters through decomposition is Tensor Train Decomposition.
This approach, on the other hand, is $N$-dimensional, which means that it can be used to decompose a matrix of any number of dimensions.
Also, the algorithm is designed in such a way that there is no option to assure that in all layers there is the same proportion of trainable parameters.
Tucker decomposition~\cite{tucker1966some} is also $N$-dimensional.
However, one of the factors of Tucker decomposition contains the number of parameters exponential in~$N$.

% \begin{figure}[h]  % TODO: not a screenshot
%     \centering
%     \includegraphics[width=0.5\textwidth]{images/conv_decomposition/tt_decomposition_illustration.png}
%     \caption{Tensor Train Decomposition.}
%     \label{fig:tt_decomposition_illustration}
% \end{figure}

To reduce the number of parameters in trainable weight matrices of convolutional layers, we propose to represent each weight matrix in Conv2d as a combination of trainable \emph{basis convolutions} (\emph{basis convs}) and all the other convolutions which are just linear combinations of the basis ones (and the decomposition coefficients are trainable).
This may be achieved with the help of QR or SVD decomposition, applied to the weight matrix reshaped to $2$D.

\paragraph{Conv2d Weight Matrix Decomposition}
Let us say, that the shape of the input of the convolution layer is $B\hmm\times C_{in} \hmm\times W \hmm\times H$, where $B$ is the number of batches, $C_{in}$ is the number of input channels, and $W$ and $H$ are the width and height respectively.
The shape of the output would be $B \hmm\times C_{out} \hmm\times W \hmm\times H$, where $C_{out}$ is the number of output channels.
The shape of the convolution layer weight matrix should be $C_{out} \hmm\times C_{in} \hmm\times K \hmm\times K$, where $K$ is the kernel size.
However, if we use basis convolutions as a decomposition method, the shape of the convolution layer's weight is going to be $C_{out}' \hmm\times C_{in} \hmm\times K \hmm\times K$, where $C_{out}'$ is the number of basis convolutions and $C_{out}' \hmm\leq C_{out}$.
Matrix decomposition goes at a cost of additional linear operation: convolution of the weight matrix with a matrix of shape $C_{out}' \hmm\times C_{out}$ to restore all the $C_{out}$ convolutions.

First, let us make sure, that matrix decomposition allows to reduce the number of trainable parameters.
Indeed, without matrix decomposition, we have $C_{out} \hmm\cdot C_{in} \hmm\cdot K \hmm\cdot K$ parameters in the weight matrix.
On the other hand, with matrix decomposition, we have a smaller weight matrix and the matrix of coefficients which allows us to restore the remaining convolutions using $C_{out}'$ basis convolutions.
So, the number of parameters equals $C_{out}' \hmm\cdot C_{in} \hmm\cdot K \hmm\cdot K \hmm+ C_{out}' \hmm\cdot C_{out}$.
Subtracting the two values representing numbers of parameters, we can see that decomposition allows using fewer parameters as long as
$
  C_{out}' \hmm< C_{out} \left[ 1 \hmm- \frac{1}{C_{in} / C_{out} \hmm\cdot K \hmm\cdot K \hmm+ 1} \right] \hmm\approx C_{out}
$,
which is always the case.

So, matrix decomposition allows to reduce the number of trainable parameters, but does it provide an increase in training speed?
Let us take some convolutional layer and denote by $N^0_f$ the number of operations in this layer during the forward pass.
We also denote by $N'_f$ number of operations in the same layer when matrix decomposition is applied.
So, without basis convolutions, we get
$
  N^0_f \hmm= K \hmm\cdot K \hmm\cdot W \hmm\cdot H \hmm\cdot C_{in} \hmm\cdot C_{out}
$.

With basis convolutions, $N'_f$ consists of two terms.
The first one corresponds to the restoration of the whole weight matrix with the help of the smaller weight matrix and the matrix of coefficients.
And the second one is just the same forward pass with the whole weight matrix.
So
$
 N'_f \hmm= C_{out} \hmm\cdot C_{out}' \hmm\cdot C_{in} \hmm\cdot K \hmm\cdot K \hmm+ K \hmm\cdot K \hmm\cdot W \hmm\cdot H \hmm\cdot C_{in} \hmm\cdot C_{out}
$.
Apparently, $N'_f \hmm> N^0_f$ which does not faster training.

\paragraph{Decomposition of Weight Matrix, Composition of the Forward's Output}
As we proved that we can use matrix decomposition to reduce the number of trainable parameters without a significant drop in quality albeit with no reduction of training time, now we are going to concentrate on speeding up the computations with basis convolutions.

\begin{figure}[h]  % TODO: svg, not png
    \centering
    \includegraphics[width=0.8\textwidth]{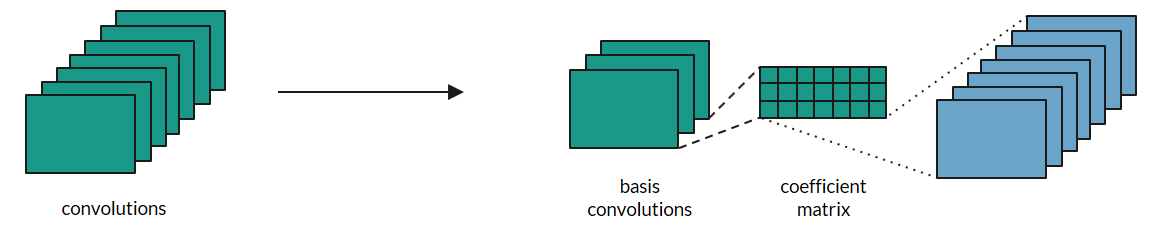}
    \caption{Basis convolutions for the forward's output modification.}
    \label{fig:basis_convs_idea2}
\end{figure}

The main idea is to try to apply basis convolutions decomposition not for the weight matrix replacement (before forward pass though the Conv2d module), but \emph{after} the actual forward pass computed with basis convolutions.
We are going to use reduced weight matrices in Conv2d modules (consisting only of basis convolutions), compute forward, and only then apply decomposition coefficient matrix to compute the outputs of all convolutions knowing the outputs just for the basis ones.
So, we partly delegate forward pass computation to the basis convolutions decomposition (see Fig.~\ref{fig:basis_convs_idea2}).
Hypothetically, this should allow us to get the desired acceleration in training.

Let us denote by $N''_f$ the number of operations in some convolution layer during the forward pass in the described scenario.
This may be represented as a sum of, first, the number of operations required for the forward pass with a smaller weight matrix consisting of basis convolutions.
Second, the number of operations required for restoration of the rest of the convolutions' output.
So, to estimate $N''_f$ we have the following:
$
 N''_f \hmm= K \hmm\cdot K \hmm\cdot W \hmm\cdot H \hmm\cdot C_{in} \hmm\cdot C_{out}' \hmm+ C_{out} \hmm\cdot C_{out}' \hmm\cdot W \hmm\cdot H
$.
Calculating the difference $N''_f \hmm- N^0_f$, we get that forward pass will require fewer operations as long as
$
  C_{out}' \hmm< C_{out} \hmm\cdot \left(1 \hmm+ \frac{C_{out}}{C_{in} \hmm\cdot K \hmm\cdot K}\right)^{-1}
$.

So, forward acceleration is possible just for relatively small values $C_{out}'$.

As long as we are aiming at the acceleration of the training process as a whole, let us also estimate the difference in the number of operations required for the \emph{backward pass}.
Backward pass for the given convolutional layer may be estimated as the number of scalar values in the input which take part in generating one scalar value of the convolution layer's output.
That is if we do not use basis convolutions, each output scalar depends on $K \hmm\cdot K \hmm\cdot C_{in}$ input scalars.
And the number of such dependencies in the whole output tensor is
$
  N^0_{b} \hmm= W \hmm\cdot H \hmm\cdot C_{out} \hmm\cdot K \hmm\cdot K \hmm\cdot C_{in}
$.

\begin{figure}[H]
    \centering
    \includegraphics[width=0.8\textwidth]{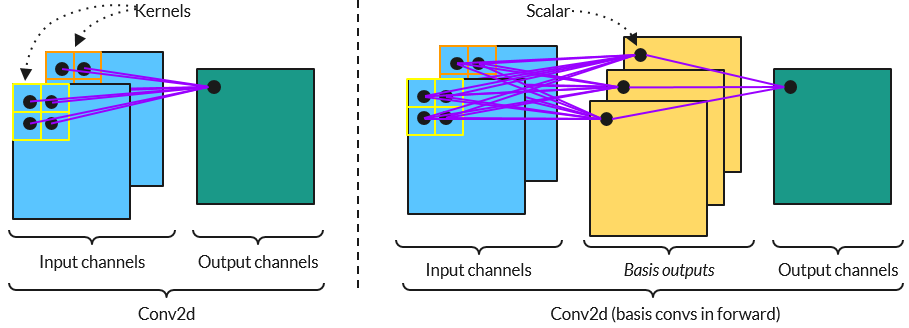}
    \caption{Dependencies of one output scalar on input scalars for Conv2d layer. Forward method's output modification with basis convs speeds up forward pass but slows down backward pass.}
    \label{fig:decomposition-for-forward-why-bad}
\end{figure}

If we now look at the corresponding value of operations $N''_{b}$ in the case of basis convolutions, we will find out that it actually increases compared to $N^0_{b}$ (see Fig.~\ref{fig:decomposition-for-forward-why-bad}).

Indeed, one scalar output requires $C_{out}' \hmm\cdot K \hmm\cdot K \hmm\cdot C_{in}$ scalar inputs.
And the whole output tensor requires
$
  N''_{b} \hmm= W \hmm\cdot H \hmm\cdot C_{out} \hmm\cdot C_{out}' \hmm\cdot K \hmm\cdot K \hmm\cdot C_{in}
$
operations, which is apparently bigger than $N^0_{b}$.

So, basis convolutions applied on the stage of the forward method's output calculation speeds up the forward pass but at the same time slows down the backward pass.
Let us estimate the difference
$
  \Delta'' \hmm= (N''_f \hmm+ N''_b) \hmm- (N^0_f \hmm+ N^0_b)
$
between total numbers of operations required for both forward and backward passes for some fixed convolution layer.
We get
$
    \Delta'' \hmm= K \hmm\cdot K \hmm\cdot W \hmm\cdot H \hmm\cdot C_{in} \hmm\cdot C_{out} \hmm\cdot \left\{
      C_{out}' \hmm+ \frac{C_{out}'}{C_{out}} \hmm+ \frac{C_{out}'}{K \hmm\cdot K \hmm\cdot C_{in}} \hmm- 2
    \right\}
$.
We achieve acceleration in training as long as $\Delta'' \hmm< 0$.
That is, as long as
$
  C_{out}' \hmm< 2 \hmm\cdot \left(1 \hmm+ \frac{1}{C_{out}} \hmm+ \frac{1}{K \hmm\cdot K \hmm\cdot C_{in}} \right)^{-1}
  \hmm\approx 2
$.

In other words, basis convolutions provide acceleration as long as $C_{out}' \hmm\leq 2$.
Obviously, such a low number of basis convolutions is going to degrade the final model quality.

% The explanation for the failure to accelerate training may be the following.
% Though the number of computations on the forward pass becomes lower, the backward pass becomes more complicated (see Fig.~\ref{fig:decomposition-for-forward-why-bad}).
% Which prevents the model from training faster.

\begin{figure}[h]
    \centering
    \includegraphics[width=0.8\textwidth]{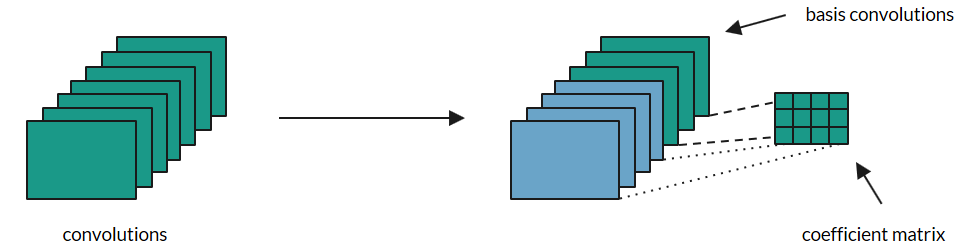}
    \caption{Some basis convolutions' output stays unchanged as the subset of the layer's resulting output.}
    \label{fig:basis_convs_idea3}
\end{figure}

\paragraph{Reducing the Size of the Coefficient Matrix for Acceleration}
To facilitate both forward and backward passes (and finally to achieve the desired increase in training speed), it is proposed to reduce the size of the coefficient matrix, which helped to restore all convolutions with the help of basis ones.
It can be achieved, for example, by keeping basis convolutions unchanged as the ones of the resulting convolutions (see Fig.~\ref{fig:basis_convs_idea3}).

To get some estimation of the possibility of acceleration, let us consider the following scenario.
Basis convolutions $C_{out}'$ are used for forward computation.
So, we get the output of dimension $B \hmm\times C_{out}' \hmm\times W \hmm\times H$.
Previously, to restore the outputs of the $C_{out}$ convolutions we used the coefficient matrix of shape $C_{out} \hmm\times C_{out}'$.
Now, we are going to calculate $(1 \hmm- \beta) \hmm\cdot C_{out}$ convolutions just as copies from $C_{out}'$, where $0 \hmm\leq \beta \hmm\leq 1$.
And each of the rest $\beta C_{out}$ convolutions we are going to calculate as a weighted sum of just two convolutions from $C_{out}'$.

So, for the number operations $N'''_f$ required for the forward pass we get
$
  N'''_f \hmm= K \hmm\cdot K \hmm\cdot W \hmm\cdot H \hmm\cdot C_{in} \hmm\cdot C_{out}'
  \hmm+ \bigl((1 \hmm- \beta) \hmm\cdot C_{out}\bigr) \hmm\cdot 1 \hmm\cdot W \hmm\cdot H
  \hmm+ \bigl(\beta C_{out}\bigr) \hmm\cdot 2 \hmm\cdot W \hmm\cdot H
$.
And the number of backward operations $N'''_b$ appears as
$
  N'''_b \hmm= W \hmm\cdot H \hmm\cdot \bigl((1 \hmm- \beta) \hmm\cdot C_{out}\bigr) \hmm\cdot 1 \hmm\cdot K \hmm\cdot K \hmm\cdot C_{in}
   \hmm+ W \hmm\cdot H \hmm\cdot \bigl(\beta \hmm\cdot C_{out}\bigr) \hmm\cdot 2 \hmm\cdot K \hmm\cdot K \cdot C_{in}
$.

To get acceleration, we want the total difference
$
  \Delta''' \hmm= (N'''_f \hmm+ N'''_b) \hmm- (N^0_f \hmm+ N^0_b)
$
to be lower than zero.
If we put $C_{out}' \hmm\equiv \alpha C_{out}$, where $0 \hmm< \alpha \hmm\leq 1$, the condition $\Delta''' \hmm< 0$ results in
$
  \beta \hmm\cdot \left\{1 \hmm+ \frac{1}{K \hmm\cdot K \hmm\cdot C_{in}}\right\} \hmm+ \alpha \hmm< 1 \hmm- \frac{1}{K \hmm\cdot K \hmm\cdot C_{in}}
$.
For the estimation purposes, it can be simplified just as $\alpha + \beta < 1$ which is feasible and, hypothetically, should not lead to the significant drop in the final model quality.
So, acceleration is possible as long as sum $\alpha$ and $\beta$ is lower than $1$, where $\alpha$ is the fraction of basis convolutions, and $\beta$ is the fraction of those ones, whose output restoration requires two basis convolutions, and the outputs of the rest $(1 \hmm- \beta)C_{out}$ convolutions are just the outputs of some of the $C_{out}'$ basis ones.

% TODO: про время вычислений; показать выигрыш

% Training pipeline with use of basis convolutions is the following.
% First, basis convolutions are trained for the first several epochs.
% Then, these convolutions are used for model initialization.
% From this point onward, the model is trained without basis convolutions.

% \begin{figure}[h]
%     \centering
%     \includegraphics[width=0.5\textwidth]{images/basis_convs/basis_convs_pipeline.png}
%     \caption{Scheme of the basis convolutions approach for initialization.}
%     \label{fig:basis_convs_pipeline}
% \end{figure}

\subsection{Layer-Wise Selective Application of Basis Convolutions}

% \paragraph{О чём.}
% Описание способа поиска оптимальной конфигурации для обучения со свёртками, в основе которого лежит идея из \citep{cai2020zeroq}.

% \paragraph{Что будет.}
% \begin{itemize}
%     \item Гипотеза о разной чувствительности слоёв сети к применению базисных свёрток.
%     \item Небольшой подтверждающий гипотезу эксперимент (обучение со свёртками только в одном слое: в первом, посередине, и в последнем): короткая постановка, без подробностей, плюс графики Valid accuracy. 
% \end{itemize}

% \bigskip

\paragraph{Sensitivity of Layers to Basis Convolutions}
So, we proved that it is possible to achieve faster training using basis convolutions.
However, the quality may appear worse than the baseline model's.
In this section, it is proposed to apply basis convolutions \emph{selectively}.
That is, we are going to propose an algorithm for finding those layers in the model, which are less \emph{sensitive} to training with basis convolutions.
If we manage to identify those layers, we can train all other, more sensitive layers, as usual.
So, hypothetically, selective application of basis convolutions should allow us to train the model faster (although not so fast as when all modules are trained with basis convolutions) and without a drop in model quality.

Summing up, the idea of the sensitivity of modules to basis convolutions can be formulated as follows:
\begin{itemize}
    \item Different layers of the model have different sensitivity to training with basis convolutions.
    \item If we apply basis convolutions only to those layers which are not very sensitive, we may get faster training without a drop in quality.
\end{itemize}

% To find these less sensitive modules, we formulate two hypotheses:
% \begin{itemize}
%     \item The higher the correlation between convolutions, the less layer sensitivity is.
%     \item The deeper the layer is located in the model (the larger the layer), the less layer sensitivity is.
% \end{itemize}

Next, we are going to analyze the sensitivity of the convolutional model's layers using the proposed hypotheses.

\paragraph{Dependence Between Layer Choice and the Resulting Model Quality}
Here, we want to find out, whether or not there is any difference if we apply basis convolutions to different modules.
To do this, we are going to conduct the following experiment: we are going to train two ResNet18 models.
The first one (the ``light'' model) is going to be trained with basis convolutions applied to only the first modules of the model (layers number $1$, $2$, $3$, $4$, $5$).
The second one (the ``heavy'' model), on the other hand, is going to be trained with basis convolutions applied to only the last modules of the model (layers number $16$, $17$, $18$, $19$, $20$).
Hypothetically, these two groups of modules should show the different resulting quality if there is such a notion as the sensitivity of a module to basis convolutions.
In the Appendix~\ref{sec:best-combinations-with-convs} there is a numbered list of $2$D convolutional layers of the ResNet18 model.

The experiment setting is the following.
We train ResNet18 model on CIFAR-10 dataset with cosine annealing scheduler with $T_{max} \hmm= 90$ and starting learning rate equal to $0.1$.
During training, basis convolutions are used in a group of layers: ``light'' or ``heavy''.

The results are shown in Fig.~\ref{fig:basis_convs_light_vs_heavy_accuracies}.
It can be seen that some layers are indeed more sensitive to basis convolutions than the other ones.
It means that we need to find those layers where we can use basis convolutions without a drop in the final model quality.

% \begin{figure}[h]
%     \centering
%     \begin{subfigure}[h]{0.5\columnwidth}
%         \centering
%         \includegraphics[width=\textwidth]{images/basis_convs/basis_convs_light_vs_heavy_accuracies.png}
%     \end{subfigure}
    
%     \caption{Validation accuracy for models where basis convolutions are applied in different layer groups (``light'' and ``heavy'').}
%     \label{fig:basis_convs_light_vs_heavy_accuracies}
% \end{figure}

% \begin{figure}[h]
%     \centering
%     \begin{subfigure}[h]{0.45\columnwidth}
%         \centering
%         \includegraphics[width=\textwidth]{images/basis_convs/basis_convs_light_vs_heavy_relative_sizes.png}
%     \end{subfigure}
%     ~
%     \begin{subfigure}[h]{0.45\columnwidth}
%         \centering
%         \includegraphics[width=\textwidth]{images/basis_convs/basis_convs_light_vs_heavy_absolute_sizes.png}
%     \end{subfigure}
    
%     \caption{Left two plots: difference in size and training time for ``light'' and ``heavy'' combinations comparing to baseline. Right two plots: absolute values of models' size and training time.}
%     \label{fig:basis_convs_light_vs_heavy_sizes}
% \end{figure}

\paragraph{Search for Layer Subset}
If we want to find an optimal combination of layers where we can apply basis convolutions without a drop in quality, we face the following problem: we should train an exponential number of models.
For example, if there are $L$ convolutional layers in the model, the number of experiments required to find an optimal layer combination, which provides acceleration with no or almost no drop in quality, would be $2^L$.

Obviously, for deep neural networks, training such a big number of models in a reasonable time is just not feasible.
We propose the following solution: to train a range of models where basis convolutions are applied in \emph{one layer only}.
So, the number of trained models will be linear with the number of layers, not exponential.
This range of models provides us with the estimates of \emph{sensitivity} of each particular layer to basis convolutions.
We can estimate layer sensitivity as a drop in final model quality if we apply basis convolutions in the specified layer.
Using this range of models, we can combine obtained sensitivities to estimate the sensitivity of the whole \emph{layer groups} to basis convolutions and thus find an optimal layer subset.

The presented idea is inspired by the work~\cite{cai2020zeroq} where the authors deal with the quantization of neural networks.
This is somewhat similar to the use of basis convolutions.
In quantization, a model is trained with its parameters stored in lower bitwidths.
With basis convolutions, a model is trained with just less trainable parameters.

\paragraph{Sensitivity of Compound Model to Basis Convolutions}
Here we are going to combine the obtained one-layer estimates to calculate the sensitivity of layer groups to basis convolutions.
We propose the following algorithm.
For each layer, there are two possibilities: if it is trained with or without basis convolutions.
If a layer is selected as the one with basis convolutions, we take training time, model size, and accuracy of the model trained as in the previous section, where basis convolutions were applied in this particular layer only.
We compute an ``accuracy drop'' corresponding to the module which is a difference between the baseline's accuracy and accuracy of the model which has this layer decomposed with basis convolutions.
If a layer is selected as trained without basis convolutions, we take training time, model size, and accuracy of the baseline ResNet18 model.
For each layer combination (where some layers are selected as trained with basis convolutions and others are not) we average training time, model size, and valid accuracy drop estimates over the number of layers.
Thus, we get a set of module combinations with estimates of the metrics corresponding to the model as a whole.

\paragraph{Application of Basis Convolutions}

We see two main scenarios of how basis convolutions can be used to increase model training or inference time.

First, basis convolutions can be trained in the found least sensitive layers from start to finish.
This would allow getting a boost in training time, inference time.

Second, we can train basis convolutions just during some first epochs of training, then compose the weight matrices of original size and proceed with the training till the end without basis convolutions.
We call such experiment setting as training with ``skip'', where the skip is a start period of time where basis convolutions are used.
This approach still allows to reduce training time (albeit the time gain is going to be lower than in the previous approach) but the final model quality is supposed to be higher.

In the experiments, we are going to use the second approach.
The whole training will be split into two sections.
The first one is training with basis convolutions applied in the least sensitive layers.
The second one is training till the end without basis convolutions.

\section{Experiments}

\subsection{General Experiment Setting}
\label{sec:general_experiment_setting}

For the experiments, we use ResNet18 model~\cite{he2016deep}, which is a convolutional network $18$ layers deep and is used for image classification tasks.
For training and validation of the model, we use CIFAR-10 dataset~\citep{krizhevsky2009learning}, which is a dataset consisting of $60000$ colored images of size $32 \hmm\times 32$ split into $10$ equally sized distinct classes.

The standard training pipeline is the following.
We train a model for $90$ epochs with a cosine annealing learning rate (LR) scheduler starting from LR equal to $0.1$.
Optimizer is SGD with momentum~\citep{qian1999momentum} equal to $0.9$ and weight decay\citep{loshchilov2017decoupled} equal to $0.0005$.
During training, all images are resized to $256$ and then cropped (randomly) to $224$.
During validation, all images are resized to $224$.

% TODO: Иллюстрация с архитектурой модели.
% TODO: CIFAR100.

% TODO: Иллюстрация с несколькими картинками из датасетов (?)

\subsection{Skip Setting}

% \paragraph{О чём.}
% Обучение моделей со Skip-ом, с тем чтобы большой LR не портил инициализацию.

% TODO: Иллюстрация с графиками Valid accuracy при инициализации обученным чекпоинтом при разных Skip-ах.
As basis convolutions provide faster training but at a cost of the final model quality degradation, we are going to use basis convolutions during only the \emph{first part} of the training process.
This way, we may achieve acceleration without degrading the quality of the model.

For training with basis convolutions, we use the setting which we call \emph{training with skip}.
The idea is to train the model with basis convolutions for some number of epochs, then use matrix decomposition to restore weight matrices of original size and proceed with the training till the end.
We call the period of training with basis convolutions as \emph{skip}, because this is the part of the general training setting which is ``skipped'', replaced by the modified one.
So, after the skip, only model architecture undergoes changes.
Learning rate history stays untouched.

\subsection{Opportunity to Achieve Faster Training with Lower Model Size}

% \paragraph{О чём.}
% Сравнение ускорений.
% Сравнение итоговых значений Valid accuracy.
% Сравнение размеров модели.

% \paragraph{Постановка.}
% Обучение сетей ResNet (возможно, только одной сети) с использованием всех рассмотренных методов для уменьшения числа параметров сети: TTD, \todo{Tucker}, Basis convs 1 (замена weight матрицы), Basis convs 2 (модификация forward метода).

% \paragraph{Что будет.}
% \begin{itemize}
%     \item Графики Valid accuracy в зависимости от времени.
%     \item Bar chart (или pie chart), иллюстрирующий разницу в размере модели.
%     \item Таблица со значениями training time (в процентах от обычной сети ResNet).
%     \item Таблица со значениями model size (в процентах от обычной сети ResNet).
% \end{itemize}

% \paragraph{Итог.}
% Базисные свёртки позволяют ускорять обучение и уменьшать размер модели.
% Но качество сильно просаживается, если учить со свёртками.

% \bigskip

In this section, we want to find if it is at all possible to accelerate training if we apply matrix decomposition to the weight matrices of the model's convolutional layers.

\paragraph{Sanity Check}
The purpose of this section is to check if a model can achieve high quality when using basis convolutions for initialization (QR and SVD decomposition).

We train the model till the end with the standard training pipeline.
After that, we extract basis convolutions from trained weight matrices.
Then we initialize a new network using the extracted basis convolutions.
And train it from $skip \hmm= 0.7$.
That is, we just skip $70\%$ of the training.
We also vary the number of basis convolutions used for the initialization.

The results are presented in Fig.~\ref{fig:sanity-check-qr-svd}.
It can be seen that the model recovers.
As expected, the model size is reduced.
The resulting quality, however, is comparable to one of the baseline model.
Training time remains unchanged.
So we see that basis convolutions do not necessarily degrade quality, but we also achieve no speed up, and so no positive training effect from the application of basis convolutions.

\paragraph{Reducing the Size of the Coefficient Matrix for Acceleration}
% TODO: про время вычислений; показать выигрыш
For comparison, we try different skip values.
Before the skip, basis convolutions are training.
After the skip, we return back to ordinary model architecture and initialize it using trained basis convolutions.
We also vary the number of basis convolutions ($1$ basis convolutions for all modules; and $0.1$ or $0.5$ fraction of module's convolutions).

Although there is a small increase in training speed for the first epochs, the final quality is worse compared to the baseline's (\ref{fig:basis_convs_pipeline_results}).

% \begin{figure}[h]
%     \centering
%     \includegraphics[width=0.6\textwidth]{images/basis_convs/basis_convs_pipeline_results.png}
%     \caption{Valid accuracy for ResNet18 on CIFAR10. ``Skip'' means the number of skipped epochs. ``Size'' means the number (if size is integer) or fraction (if size is float) of basis convolutions in the module of all the convolutions in this module. ``Baseline'' is the standard training pipeline. ``Random'' is random initialization.}
%     \label{fig:basis_convs_pipeline_results}
% \end{figure}

\paragraph{Conclusion}

Basis convolutions make it possible to train a model faster.
The model size is also reduced.
However, the quality drops significantly if we train the model with basis convolutions in all convolutional layers.

\subsection{Layer-Wise Selective Application of Basis Convolutions}
\label{ses:layerwise_application}

In this section, we want to get acceleration, but with no or almost no drop in the final model quality.
To do this, we are going to select the \emph{subset} of all convolutional layers where we will apply basis convolutions.

\paragraph{Sensitivity of Layers to Basis Convolutions}
Here, we apply basis convolutions in one layer of the model only.
We train a range of such models which differ in the layers where basis convolutions are applied.

The results are presented in Fig.~\ref{fig:basis_convs_per_layer_accuracy}.
It can be seen, that the first layer is the one which is the most sensitive to training with basis convolutions.
Green bars on the time chart correspond to the layers where basis convolutions led to faster training (see the ordered sequence of ResNet18 convolutional layers in the appendix~\ref{sec:best-combinations-with-convs}).

\paragraph{Sensitivity of Compound Model to Basis Convolutions}
We can plot the points on the graph.
So, the results are presented in Fig.~\ref{fig:basis_convs_clouds}.
Each point on the plot represents a model where several layers are trained with basis convolutions.
Thus, a point is defined by a \emph{combination} of layers.
Each point is associated with a valid accuracy drop (sensitivity estimate), training time estimate, and model size estimate.

Best selected layer combinations (red points on the plot in Fig.~\ref{fig:basis_convs_clouds}) are the following: $\{6, 10, 12, 14, 15, 17, 19, 20\}$, $\{4, 6, 15, 19, 20\}$, $\{12, 15, 20\}$, $\{4, 15\}$
which correspond to training time $191.5$, $192$, $192.5$, $193$ minutes respectively (ResNet18 layer sequence is presented in the Appendix~\ref{sec:best-combinations-with-convs}).

% \begin{figure}[H]
%     \centering
%     \begin{subfigure}[h]{0.45\columnwidth}
%         \centering
%         \includegraphics[width=\textwidth]{images/basis_convs/basis_convs_cloud_time.png}
%     \end{subfigure}
%     ~
%     \begin{subfigure}[h]{0.45\columnwidth}
%         \centering
%         \includegraphics[width=\textwidth]{images/basis_convs/basis_convs_cloud_size.png}
%     \end{subfigure}
    
%     \caption{Valid accuracy drop depending on training time (left) and model size (right). The green point represents a model without basis convolutions (zero accuracy drop on both plots). Valid accuracy drop is computed as ``baseline accuracy'' minus ``obtained accuracy'' (the lower the better). Red points are the selected layer combinations which show training time lower than the one corresponding to the model without basis convolutions, but whose valid accuracy drop is as low as possible.}
%     \label{fig:basis_convs_clouds}
% \end{figure}

\paragraph{Application of Basis Convolutions}
In this section, we describe the final experiment with basis convolutions.
We are going to apply basis convolutions in the found layer subsets.
Basis convolutions are trained for several first epochs (``skip'').
After training of basis convolutions, the model is reinitialized and trained till the end without basis convolutions.

Results are presented in Fig.~\ref{fig:basis_convs_finale_skip8}.
%, and Fig.~\ref{fig:basis_convs_finale_skip7}.
It can be seen that the proposed selective application of basis convolutions allowed to achieve an increase in training speed (albeit small) without a drop in valid accuracy.

\section{Conclusion}

Matrix decomposition is a way to reduce the number of model parameters.
However, it might not lead to the reduction of training time.
In this paper, we propose a new approach of matrix decomposition application to the weight matrices of convolutional layers.
We keep a smaller number of trainable convolutions (basis convolutions), and simulate the rest of the convolutions as linear combinations of the basis ones.
Importantly, to get acceleration in training we apply decomposition logic \emph{after} the forward method of the layer computed with the help of basis convolutions.

\section{Future Work}
\label{sec:future_work}

In the future, we are aiming to reduce the time needed for the search of the optimal layer combination which can be trained with basis convolutions.
Currently, it requires training of the linear number of models.
We also find it important to propose a way to select the number of basis convolutions in each layer so as to get a provable acceleration in the end.
In this paper, we proved that acceleration is possible, fixed the proportion of basis convolutions to be the same in all layers of the model, and used sensitivity analysis to find a subset of layers.
Basis convolutions applied in particular layers even increased training time.
So, we may get provable acceleration with the finer tuning of basis convolutions fraction among layers.
All experiments were conducted with the ResNet18 model on the CIFAR-10 dataset.
We find it sufficient to prove the proposed ideas.
However, in future work, we are going to use a broader range of convolutional models and datasets to investigate the impact of basis convolutions on different architectures.

\bibliography{references}

% and for biber / biblatex, use:

% \printbibliography

% supplemental material -- everything hereafter will be suppressed during
% submission time if the hidesupplement option is provided!

\appendix

\newpage

\section{Appendix 1: Plots}
\label{sec:appendix-plots}

\begin{figure}[H]
    \centering
    \begin{subfigure}[h]{0.6\columnwidth}
        \centering
        \includegraphics[width=\textwidth]{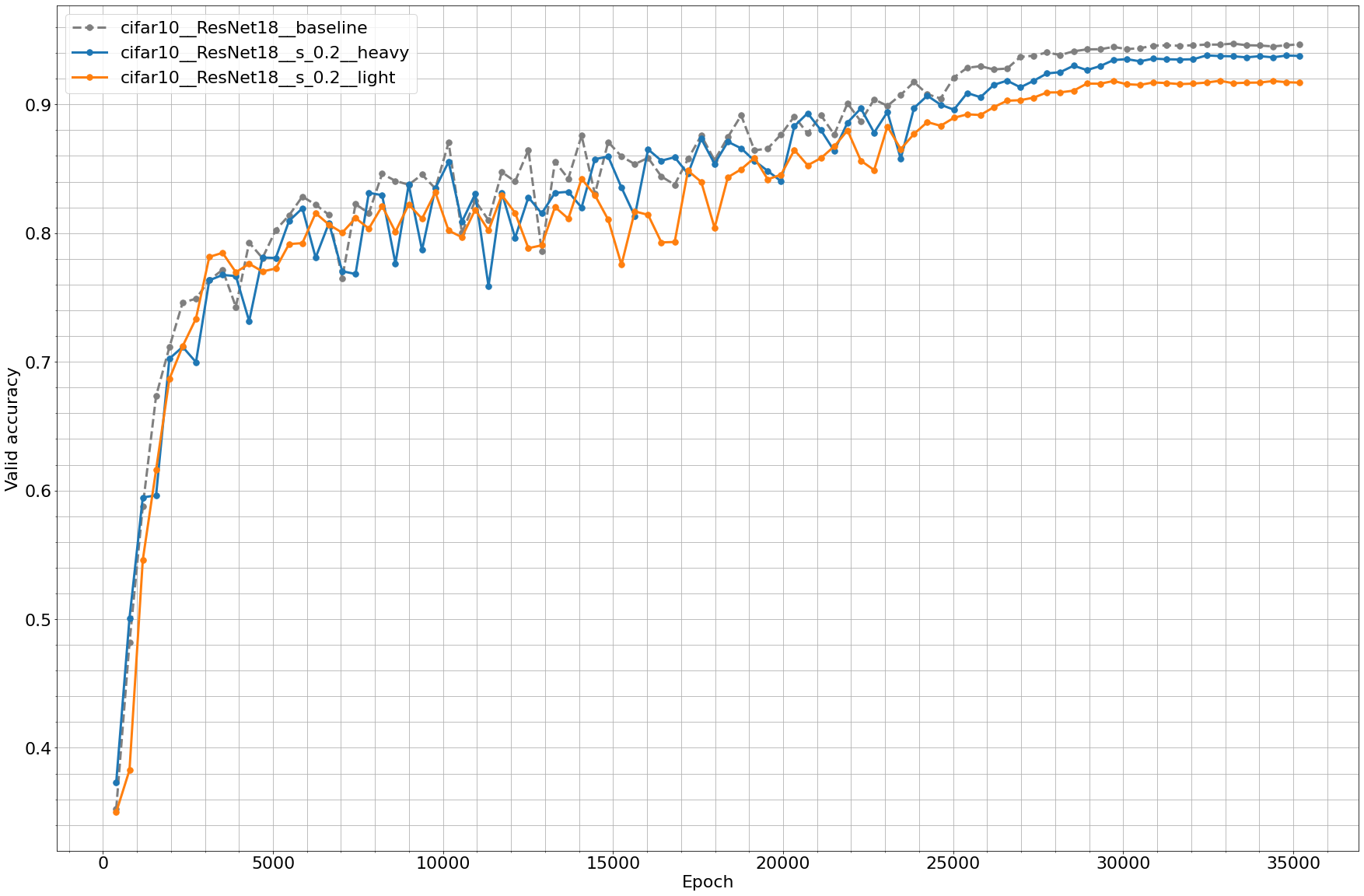}
    \end{subfigure}
    
    \caption{Validation accuracy for models where basis convolutions are applied in different layer groups (``light'' and ``heavy'').}
    \label{fig:basis_convs_light_vs_heavy_accuracies}
\end{figure}

\begin{figure}[h]
    \centering
    \begin{subfigure}[h]{0.45\columnwidth}
        \centering
        \includegraphics[width=\textwidth]{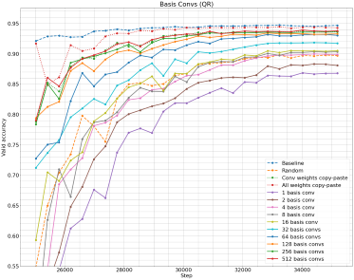}
    \end{subfigure}
    \hfill
    \begin{subfigure}[h]{0.45\columnwidth}
        \centering
        \includegraphics[width=\textwidth]{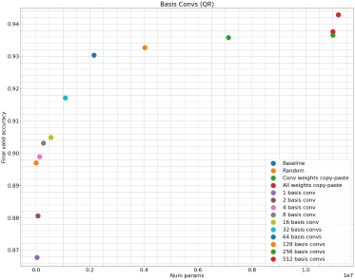}
    \end{subfigure}
    
    % \begin{subfigure}[h]{0.45\columnwidth}
    %     \centering
    %     \includegraphics[width=\textwidth]{images/conv_decomposition/sanity_check_svd.png}
    % \end{subfigure}
    % \hfill
    % \begin{subfigure}[h]{0.45\columnwidth}
    %     \centering
    %     \includegraphics[width=\textwidth]{images/conv_decomposition/sanity_check_svd2.png}
    % \end{subfigure}
    
    \caption{QR matrix decomposition sanity check experiment. Left: valid accuracy depending on train step. ``Baseline'' is random initialization from start ($skip \hmm= 0$). Only last $27$ epochs showed. ``Random'' is random initialization ($skip \hmm= 0.7$). ``Copy-paste'' is just a copy of the pretrained weights ($skip \hmm= 0.7$). Right: final accuracy depending on the number of parameters extracted from the pretrained checkpoint.}
    \label{fig:sanity-check-qr-svd}
\end{figure}

\begin{figure}[H]
    \centering
    \begin{subfigure}[h]{0.45\columnwidth}
        \centering
        \includegraphics[width=\textwidth]{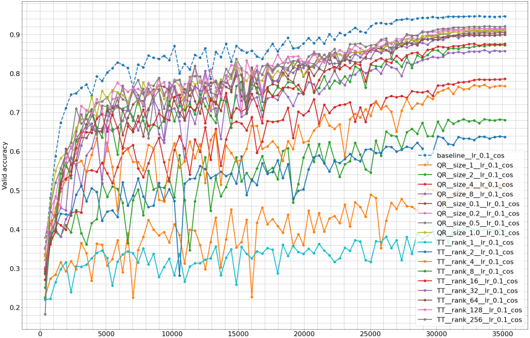}
    \end{subfigure}
    \hfill
    \begin{subfigure}[h]{0.45\columnwidth}
        \centering
        \includegraphics[width=\textwidth]{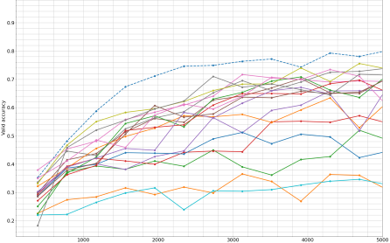}
    \end{subfigure}
    
    \caption{Training curves for models trained with standard config but with different initialization. ``Baseline'' means random initialization.}
    \label{fig:results-decomposition-weight}
\end{figure}

\begin{figure}[h]
    \centering
    \includegraphics[width=0.6\textwidth]{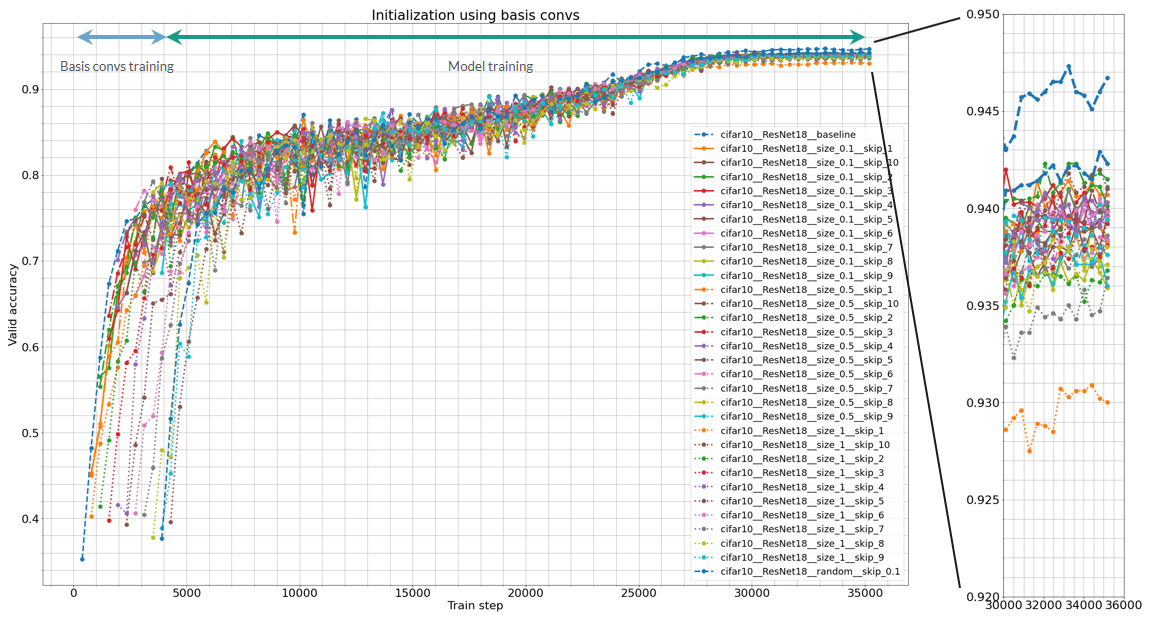}
    \caption{Valid accuracy for ResNet18 on CIFAR10. ``Skip'' means the number of skipped epochs. ``Size'' means the number (if size is integer) or fraction (if size is float) of basis convolutions in the module of all the convolutions in this module. ``Baseline'' is the standard training pipeline. ``Random'' is random initialization.}
    \label{fig:basis_convs_pipeline_results}
\end{figure}

\begin{figure}[h]
    \centering
    \begin{subfigure}[h]{0.45\columnwidth}
        \centering
        \includegraphics[width=\textwidth]{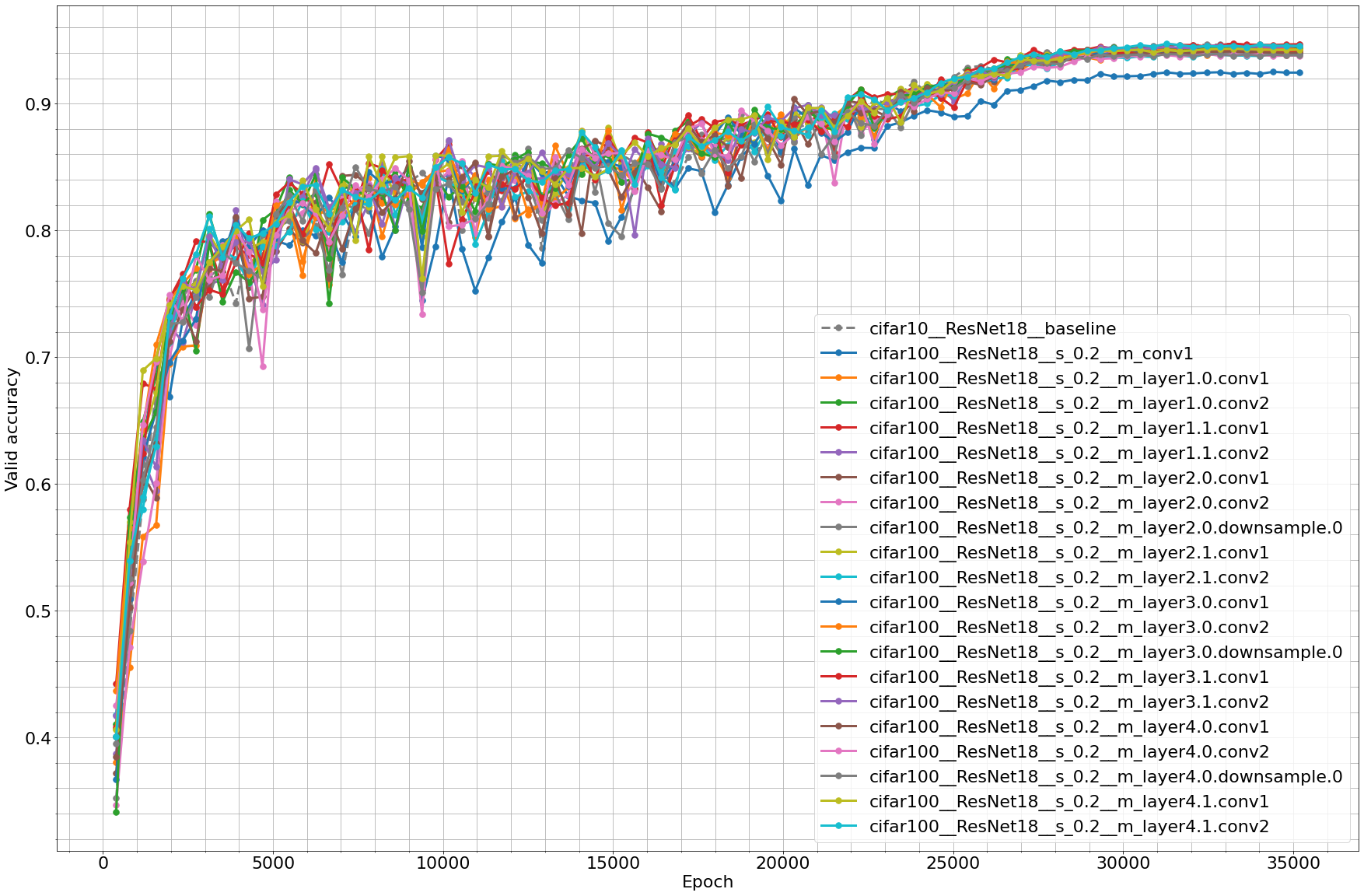}
    \end{subfigure}
    ~
    \begin{subfigure}[h]{0.08\columnwidth}
        \centering
        \includegraphics[width=\textwidth]{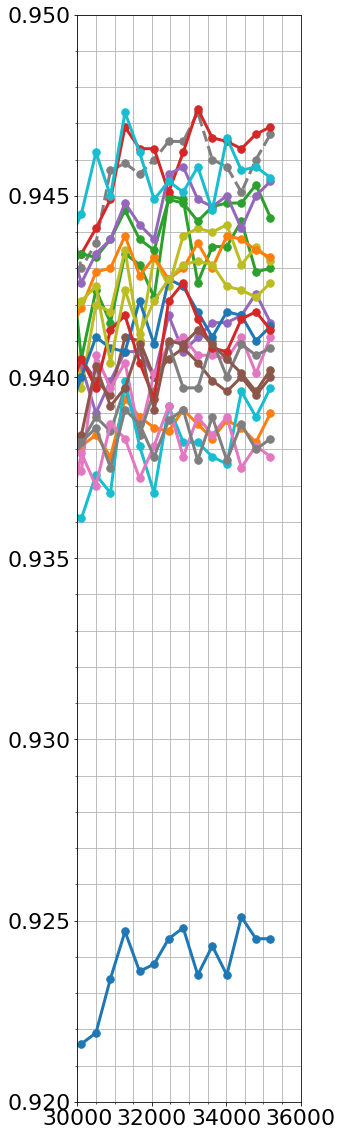}
    \end{subfigure}
    ~
    \begin{subfigure}[h]{0.3\columnwidth}
        \centering
        \includegraphics[width=\textwidth]{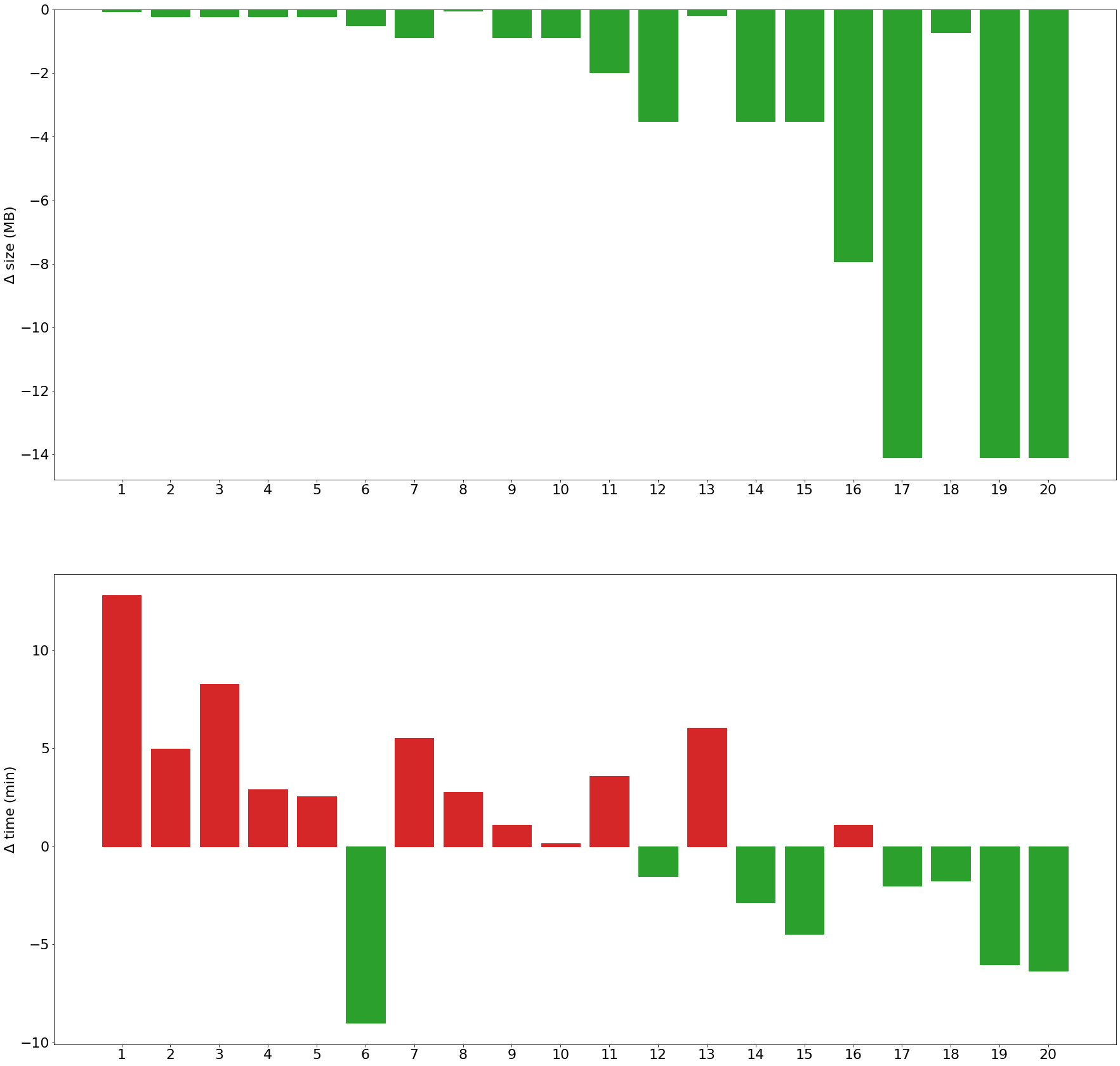}
    \end{subfigure}
    
    \caption{Left: valid accuracy for models where basis convolutions are applied in one layer only. Difference in size (upper right plot) and training time (lower right plot) comparing to baseline.}
    \label{fig:basis_convs_per_layer_accuracy}
\end{figure}

\begin{figure}[H]
    \centering
    \begin{subfigure}[h]{0.45\columnwidth}
        \centering
        \includegraphics[width=\textwidth]{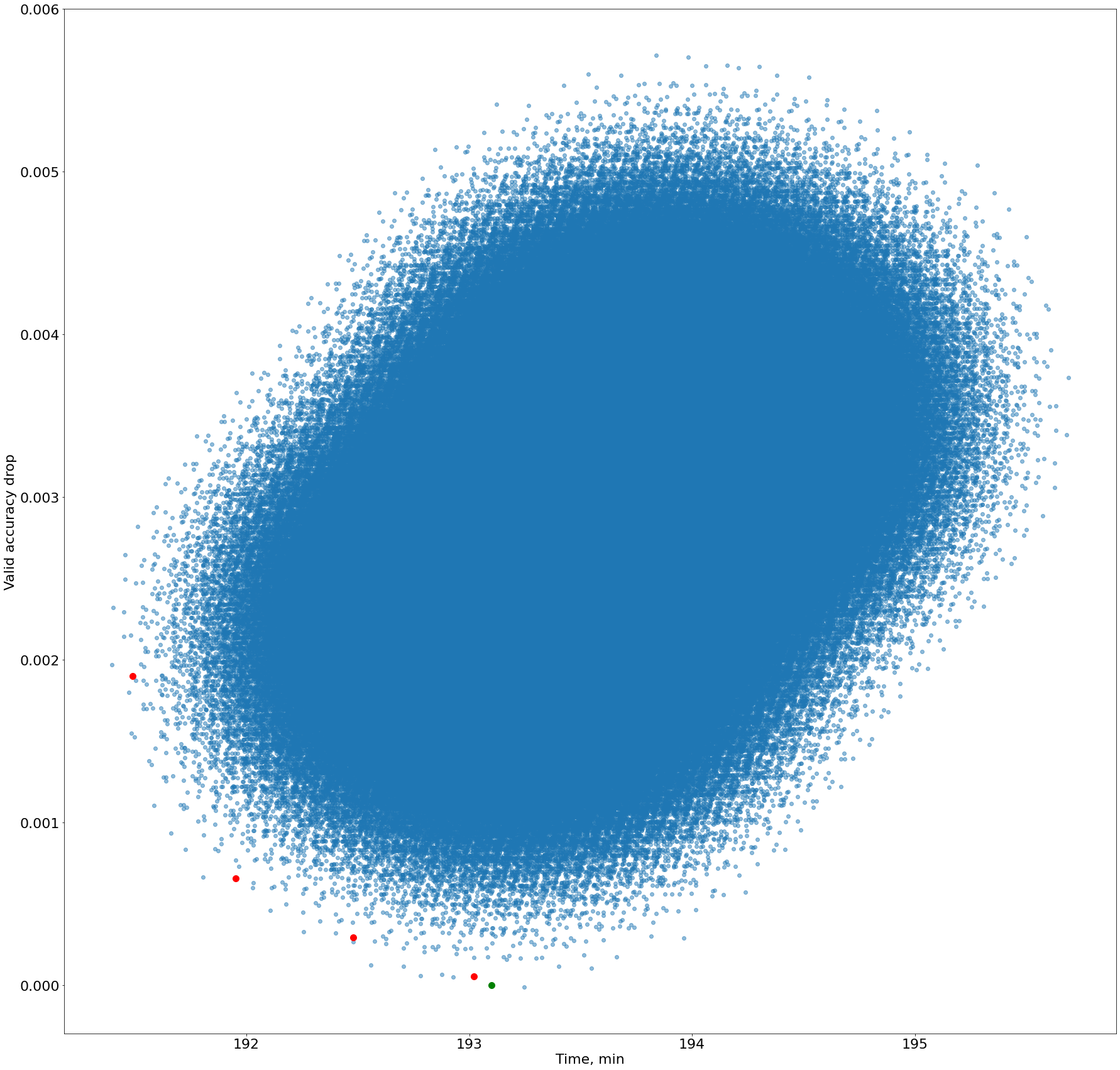}
    \end{subfigure}
    ~
    \begin{subfigure}[h]{0.45\columnwidth}
        \centering
        \includegraphics[width=\textwidth]{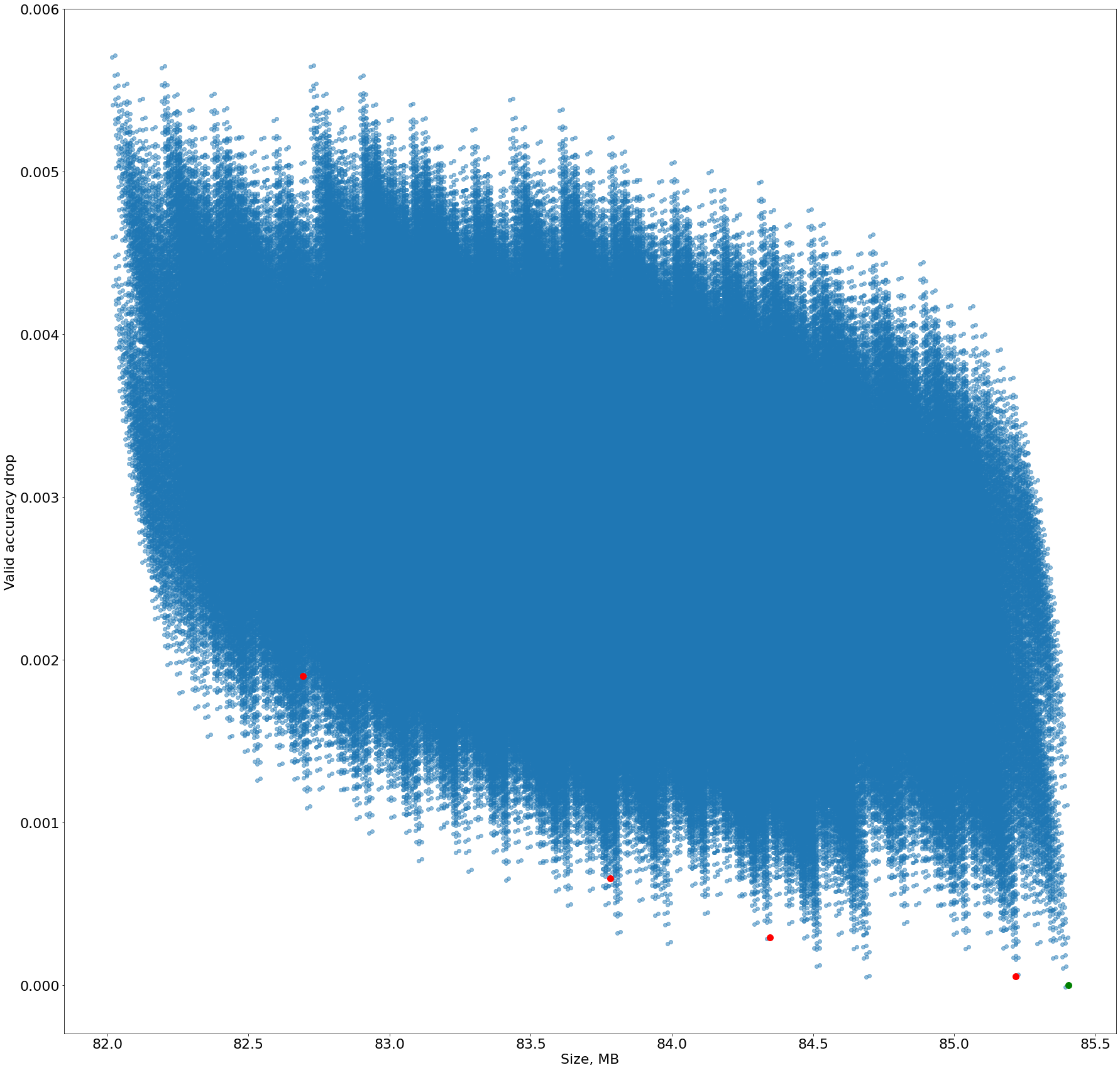}
    \end{subfigure}
    
    \caption{Valid accuracy drop depending on training time (left) and model size (right). The green point represents a model without basis convolutions (zero accuracy drop on both plots). Valid accuracy drop is computed as ``baseline accuracy'' minus ``obtained accuracy'' (the lower the better). Red points are the selected layer combinations which show training time lower than the one corresponding to the model without basis convolutions, but whose valid accuracy drop is as low as possible.}
    \label{fig:basis_convs_clouds}
\end{figure}

\begin{figure}[h]
    \centering
    \begin{subfigure}[h]{0.45\columnwidth}
        \centering
        \includegraphics[width=\textwidth]{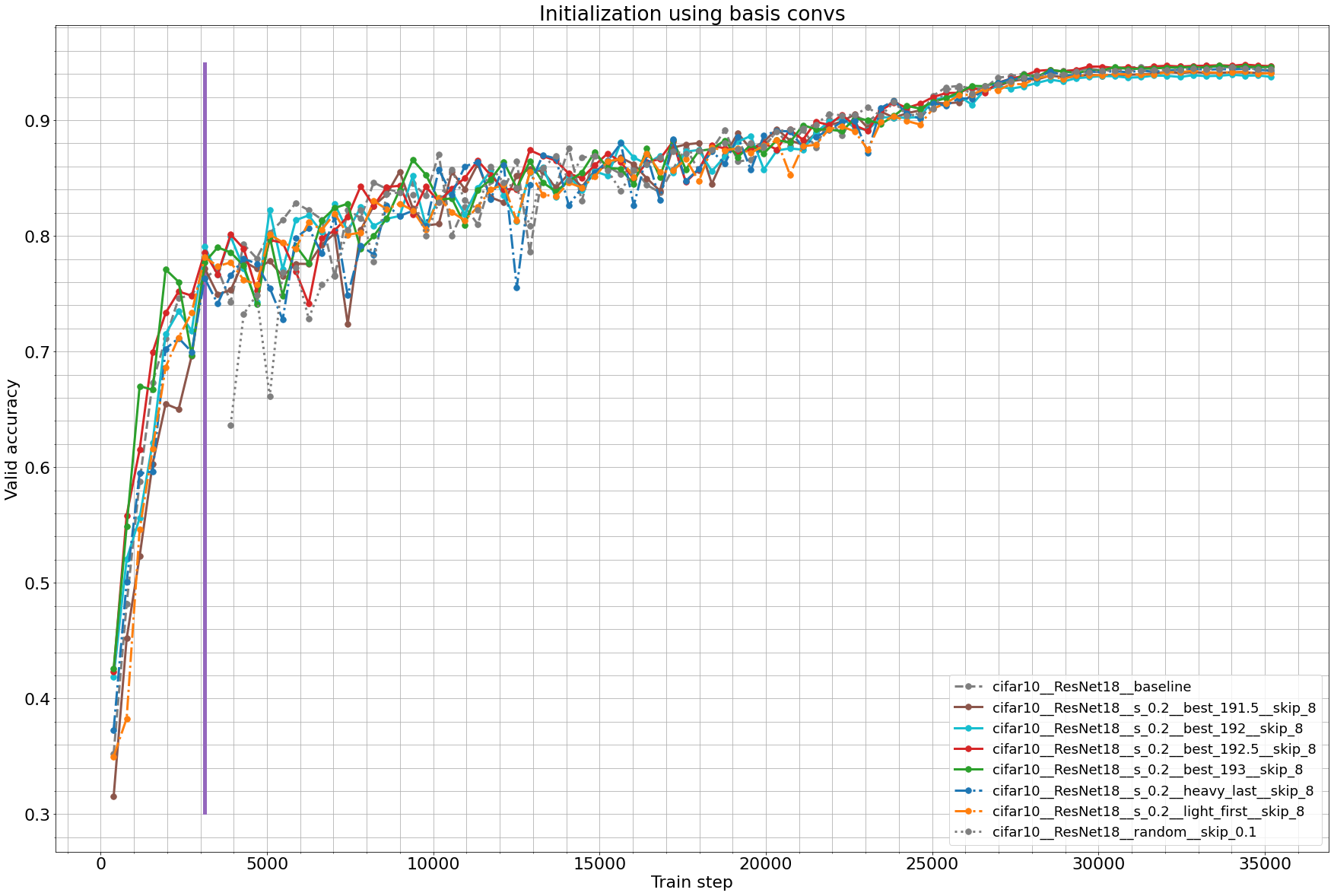}
    \end{subfigure}
    ~
    \begin{subfigure}[h]{0.08\columnwidth}
        \centering
        \includegraphics[width=\textwidth]{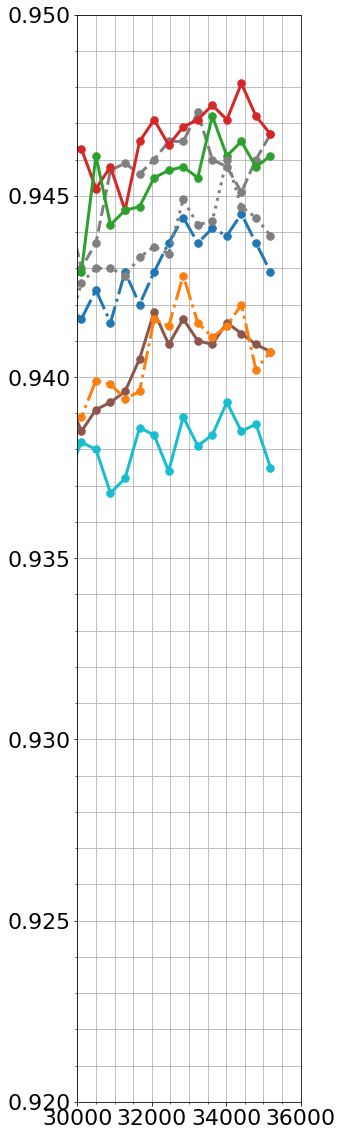}
    \end{subfigure}
    ~
    \begin{subfigure}[h]{0.3\columnwidth}
        \centering
        \includegraphics[width=\textwidth]{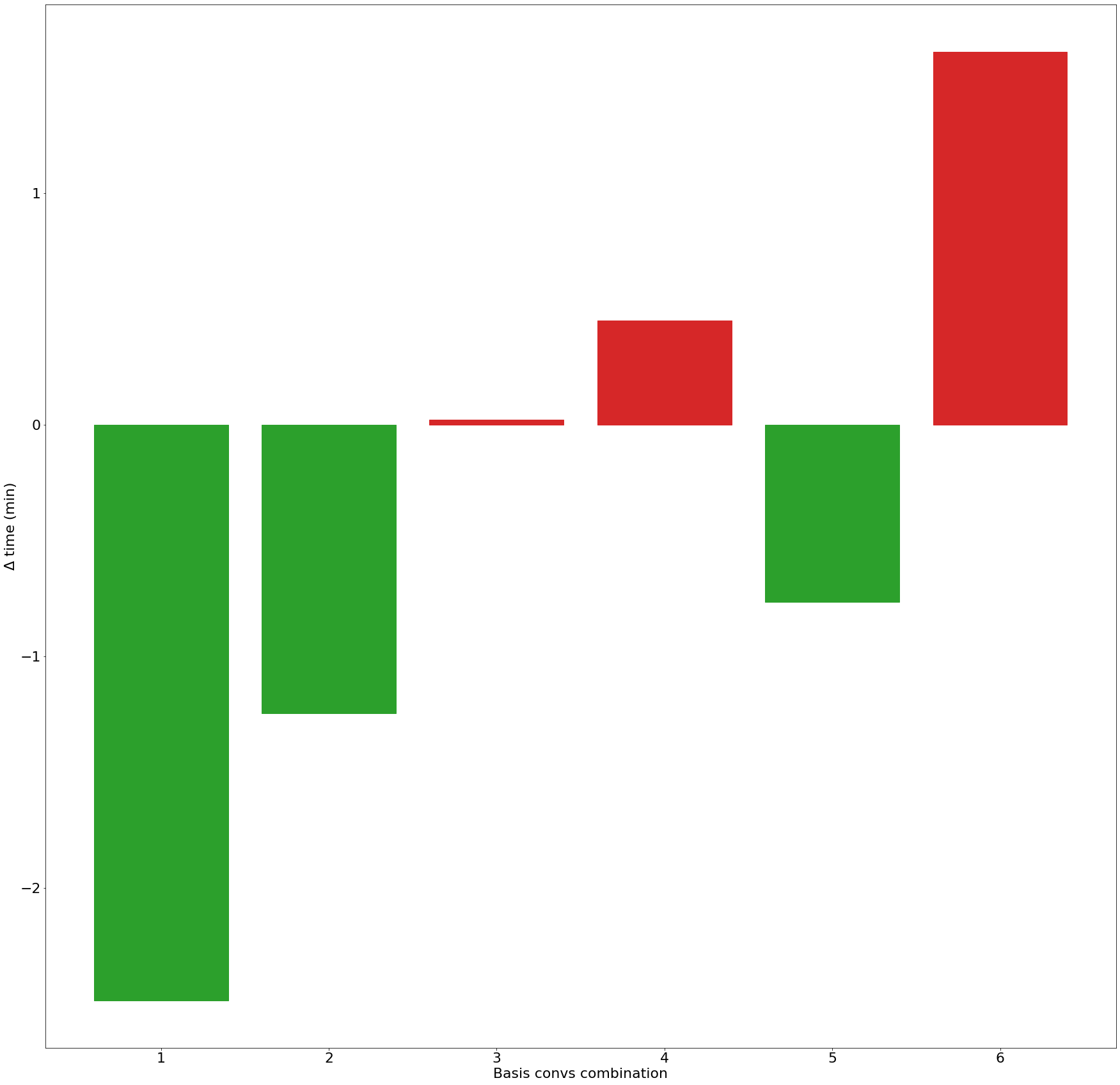}
    \end{subfigure}
    
    \caption{Valid accuracy for different combinations of layers where basis convolutions are applied (upper). The purple vertical line represents the end of training basis convolutions in the selected layers: the model is initialized, and the training process resumes from the same LR which reached the model with basis convolutions. The difference in training time compared to baseline (lower). Bars from left to right: corresponding to layer combination near $191.5$, $192$, $192.5$, $193$ minutes, and the ``heavy'' and ``light'' ones. Basis convolutions are trained for $8$ epochs.}
    \label{fig:basis_convs_finale_skip8}
\end{figure}

\newpage

\section{Appendix 2: ResNet18 Convolutional Layers}
\label{sec:best-combinations-with-convs}

ResNet18 convolution layers:
\begin{enumerate}
    \small
    \setlength\itemsep{-0.5em}
    \item conv1
    \item layer1.0.conv1
    \item layer1.0.conv2
    \item layer1.1.conv1
    \item layer1.1.conv2
    \item layer2.0.conv1
    \item layer2.0.conv2
    \item layer2.0.downsample.0
    \item layer2.1.conv1
    \item layer2.1.conv2
    \item layer3.0.conv1
    \item layer3.0.conv2
    \item layer3.0.downsample.0
    \item layer3.1.conv1
    \item layer3.1.conv2
    \item layer4.0.conv1
    \item layer4.0.conv2
    \item layer4.0.downsample.0
    \item layer4.1.conv1
    \item layer4.1.conv2
\end{enumerate}

\end{document}